%% file: ifacconf.tex
\documentclass{ifacconf}

\usepackage{graphicx}      
\usepackage{natbib}        

\usepackage{amsmath}
\usepackage{amssymb}
\usepackage{bm}
\usepackage{algorithm}
\usepackage{algpseudocode}
\begin{document}
\begin{frontmatter}

\title{An Efficient Numerical Function Optimization Framework for Constrained Nonlinear Robotic Problems\thanksref{footnoteinfo}}

\thanks[footnoteinfo]{This work was supported in part by the Robot Industry Core Technology Development Program under Grant 00416440 funded by the Korea Ministry of Trade, Industry and Energy (MOTIE).}

\author{Sait Sovukluk$^1$}
\author{Christian Ott$^{1,2}$} 

\address{$^{1}$Automation and Control Institute (ACIN), TU Wien, 1040 Vienna, Austria (e-mail: sovukluk@acin.tuwien.ac.at, christian.ott@tuwien.ac.at).}
\address{$^{2}$Institute of Robotics and Mechatronics, German Aerospace Center (DLR), 82234 Weßling, Germany.}

\begin{abstract}                
This paper presents a numerical function optimization framework designed for constrained optimization problems in robotics. The tool is designed with real-time considerations and is suitable for online trajectory and control input optimization problems. The proposed framework does not require any analytical representation of the problem and works with constrained block-box optimization functions. The method combines first-order gradient-based line search algorithms with constraint prioritization through nullspace projections onto constraint Jacobian space. The tool is implemented in C++ and provided online for community use, along with some numerical and robotic example implementations presented in the end.
\end{abstract}


\end{frontmatter}

\section{Introduction}
Numerical optimization (\cite{numericalOptim}) is one of the most fundamental backbones of robotics. Numerical optimization methods emerge to address problems that have no analytical solutions and are subject to constraints. Such problems constitute a vast variety of applications, some examples can be listed as mechanical design topology optimization for the best jumping performance (\cite{saltoRef}), behavioral optimizations to perform certain tasks with the least energy consumption (\cite{energyRef1,energyRef2}), a control input optimization that also accounts for multiple tasks, limits, contact constraints, and closed kinematic chains (\cite{sovukluk1, khatipRef}), motion planning for complex robotic systems (\cite{wholeBodyMotionRef, sovukluk2023cascaded, westervelt2018feedback}), apex-to-apex periodic behavior optimization for nonlinear and underactuated systems (\cite{sovukluk2}), path planning optimizations for autonomous devices (\cite{trajOptim}), and so on.

Convex optimization of linear(ized) robotic system problems are studied and explored extensively (\cite{numericalOptim,optimalRef}). Such problems can scale up to thousands of parameters and still be solved in real-time (\cite{osqpRef, proxqpRef, qpswiftRef}). On the other hand, nonlinear system optimization problems are less straightforward to generalize and do not scale as such due to the numerical complexities and computational costs. Such problems are addressed by nonlinear programming (\cite{nonlinearProg}) and are also covered extensively. Some well-known community-available tools, such as CasADi (\cite{casadiRef}) and IPOPT (\cite{ipopt}) are available for analytically representable nonlinear problems.

Robotic optimization problems, on the other hand, are usually not analytically representable due to the high dimensionality and nonlinearities. Let
\begin{equation} \label{systemDyn}
\bm{M}(\bm{\nu}) \dot{\bm{\nu}}
+ \bm{C}(\bm{q},\bm{\nu})\bm{\nu} + \bm{\tau}_{g}(\bm{q}) = 
\begin{bmatrix}
    \bm{0}\\
    \bm{\tau}
\end{bmatrix}
+
\bm{J}_{c}(\bm{q})^{\top} \bm{f}_{c},
\end{equation}
be a floating base robot dynamics, where $\bm{q}$ is a set of configuration variables and $\bm{\nu} = (\bm{\nu}_{b}, \bm{\nu}_{j})$ are the generalized velocity where $\bm{\nu}_{b} = (\bm{v}_{b},\bm{\omega}_{b}) \in \mathbb{R}^{6}$ is the linear and angular velocity of the floating base and $\bm{\nu}_{j} \in \mathbb{R}^{n}$ is the generalized velocity of the joints. Due to their complexity and nonlinearity, such system dynamics are usually calculated through rigid body algorithms (\cite{featherstone2014rigid}). Some community available tools are Pinocchio (\cite{pinocchioweb,carpentier2019pinocchio}), MuJoCo (\cite{todorov2012mujoco}), and RBDL (\cite{RBDL}). As a result, the analytical representation of gradients and Hessians of optimization problems that include system dynamics and kinematics such as
\begin{equation} \label{optProblemIntro}
\begin{gathered}
    \min_{\bm{x}}{f(\bm{x}, \textit{System Dynamics}(\bm{x}), \textit{Kinematics}(\bm{x}))} \\
    \text{such that} \\
    \bm{g}_{\text{eq}}(\bm{x}, \textit{System Dynamics}(\bm{x}), \textit{Kinematics}(\bm{x})) = \bm{0} \\
    \bm{g}_{\text{ineq}}(\bm{x}, \textit{System Dynamics}(\bm{x}), \textit{Kinematics}(\bm{x})) < \bm{0} \\
\end{gathered}
\end{equation}
is not possible. As numerical Hessian estimation of such nonlinear and high dimensional problems is too expensive, a simple gradient-descent based numerical optimization method that can still take account of equality and inequality constraints may become preferable for performance and real-time implementation purposes.

This paper proposes a numerical function optimization framework that is specialized for constrained nonlinear robotic problems. The method combines first-order gradient-based line search algorithms with constraint prioritization through nullspace projections of constraint Jacobian space. The framework does not require any analytic representation of the problem and works fully numerically. Hence, it can also be referred to as a black-box optimizer. As it is too costly to compute the Hessian matrix numerically, this method employs only gradient-based search algorithms. This selection results in slower convergence but reduces the iteration cost greatly. The framework also provides a set of special update routines to update the system dynamics and kinematics either every inner iteration or once per outer iteration for performance considerations. The authors also provide the precompiled C++ libraries (ENFORCpp) for community use along with multiple numerical and one three-link robotic arm optimization implementation examples:
\begin{equation*} \label{github} \tag{$\star$}
    \text{https://github.com/ssovukluk/ENFORCpp}
\end{equation*}
%
\section{Method}
\subsection{Optimization Problem}
Let
\begin{equation} \label{optProblem}
\begin{gathered}
    \min_{\bm{x}}{f(\bm{x})} \\
    \text{such that} \\
    \bm{g}_{\text{ec}}(\bm{x}) = \bm{0} \\
    \bm{g}_{\text{ic}}(\bm{x}) < \bm{0} \\
\end{gathered}
\end{equation}
be an optimization problem as described in \eqref{optProblemIntro}, where $f(\bm{x}):\mathbb{R}^{n} \rightarrow \mathbb{R}_{\geq 0}$, $\bm{g}_{\text{ec}}(\bm{x}):\mathbb{R}^{n} \rightarrow \mathbb{R}^{n_{\text{ec}}}$, and $\bm{g}_{\text{ic}}(\bm{x}):\mathbb{R}^{n} \rightarrow \mathbb{R}^{n_{\text{ic}}}$ are the cost, equality constraint, and inequality constraint functions, respectively. Furthermore, $n \in \mathbb{N}_{>0}$, $n_{ec} \in \mathbb{N}_{0}$, and $n_{ic} \in \mathbb{N}_{0}$ represent the number of optimization parameters, number of equality constraints, and number of inequality constraints, respectively.
\subsection{Nullspace Projection}
The optimization problem requires finding the parameter set that both minimizes the cost function $f(\bm{x})$ and satisfies the equality and inequality constraints. Such problems cannot be solved simply by iterating through a gradient estimation. First, the constraints should be satisfied, and then the cost function should be iterated in the allowed directions that do not disturb any constraints but still reduce the cost. Let $g_{k}$ be the $k^{th}$ active constraint function. Furthermore, let a Jacobian matrix $\bm{J}_{\text{ac}} \in \mathbb{R}^{n_{\text{ac}} \times n}$ collect gradient transpose of all active constraint functions,
\begin{equation}
    \bm{J}_{\text{ac}} = 
    \begin{bmatrix}
        \partial g_{1}/\partial\bm{x} \\[1ex]
        \partial g_{2}/\partial\bm{x} \\
        \vdots \\
        \partial g_{n_{\text{ac}}}/\partial\bm{x} \\
    \end{bmatrix},
\end{equation}
where $n_{\text{ac}}$ represent the number of the active constraints, which may differ from the total constraint number ($n_{ec}+n_{ic}$) as not all inequality constraints may be active if they are far from the boundary. Also, let
\begin{equation}
    \nabla f^{*} = \text{proj}_{N(\bm{J})} \nabla f = (\bm{I} - \bm{J}^{\top}(\bm{J}\bm{J}^{\top})^{-1}\bm{J}) \nabla f
\end{equation}
be a nullspace projection (orthogonal projection onto nullspace) of the constraint Jacobian space such that moving towards the $-\nabla f^{*}$ direction reduces the cost function without disturbing the active constraints of the problem. Such projections will be used for hierarchical optimization between the constraint and cost functions.
\subsection{Optimization Strategy}
The optimization strategy is built on a hierarchical framework. First, all constraint functions are optimized one by one, in the nullspace of previous ones such that they do not disturb the previous constraints. Then the cost function is optimized in the nullspace of all active constraints to look for a minimal solution that satisfies all active constraints.
\subsubsection{An example optimization sequence:}
Assume an optimization problem with two constraints all of which are active, i.e., $n_\text{ac} = n_\text{ec} + n_\text{ic} = 2$. A solution sequence for such a problem can be summarized as:
\begin{enumerate}
    \item Estimate $\nabla g_{1}$, gradient of the first constraint.
    \item Optimize for $g_{1}$ through $\nabla g_{1}$.
    \item Estimate $\nabla g_{2}$.
    \item Project $\nabla g_{2}$ onto the nullspace of the constraint Jacobian, i.e., $\nabla g_{2}^{*} = \text{proj}_{N(\bm{J}_1)}\nabla g_{2}$, where $\bm{J}_1 = \begin{bmatrix} \nabla g_{1} \end{bmatrix}^{\top}$.
    \item Optimize for $g_{2}$ through $\nabla g_{2}^{*}$.
    \item Estimate $\nabla f$, the cost function gradient.
    \item Project $\nabla f$ onto the nullspace of the constraint Jacobian, i.e., $\nabla f^{*} = \text{proj}_{N(\bm{J}_2)}\nabla f$, where $\bm{J}_2 = \begin{bmatrix} \nabla g_{1}, \nabla g_{2} \end{bmatrix}^{\top}$.
    \item Optimize for the cost function through $\nabla f^{*}$.
\end{enumerate}
\subsection{Equality Constraint Optimization Subroutine}
Solving for the equality constraints is a similar procedure to solving for the cost function. As described in the previous subsection, each equality constraint is solved in the nullspace of the previous equality constraint Jacobian. The details of the equality constraint optimization algorithm are provided in Alg.~\ref{alg:eqConstAlg}.
\input{eqConst}
\subsection{Inequality Constraint Optimization Subroutine}
\begin{figure}[t!]
\centerline{\includegraphics[width=\columnwidth]{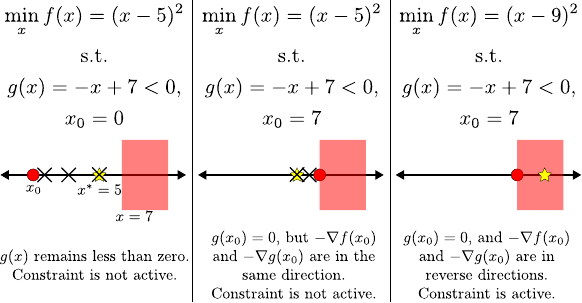}}
\caption{Inequality constraint activation conditions where the red dot, yellow star, and red region represent the initial condition, desired position, and constraint, respectively. The left problem never reaches the activation point. The middle problem starts from the zero crossing point, but the cost function gradient is already in the same direction as the constraint gradient. Hence, the cost function iteration will not exceed the constraint. The right problem also starts from the zero crossing point, and the cost function and constraint gradients are in reverse directions. The constraint is activated. The cost function gradient should be projected onto the nullspace of the constraint gradient, which results in a zero vector in this case. As a result $x$ remains at $x_{0}$.}
\label{ineqActivation}
\end{figure}

The inequality constraints are less straightforward to handle than the equality constraints due to two reasons. First, they may not always be active. Second, they are directional. As a result, these constraints should be checked continuously and activated when necessary. Furthermore, if the iteration direction aligns with the inequality constraint's gradient direction, the constraint should not be activated as any iteration in the given optimal direction would not violate the constraint. The activation conditions are summarized through an example in Fig.~\ref{ineqActivation}. A direction check between two vectors can be easily done through the dot operator. Let $\alpha \in \mathbb{R}$ be an angle between two arbitrary vectors $\bm{v}_{0}$ and $\bm{v}_{1}$, then
\begin{equation*} \label{slipOverallDynEq}
    \begin{cases}
    \enspace \bm{v}_{0} \cdot \bm{v}_{1} > 0, \enspace & 0 < \alpha < \pi/2 \\
    \enspace \bm{v}_{0} \cdot \bm{v}_{1} < 0, \enspace & \pi/2 < \alpha < \pi \\
    \enspace \bm{v}_{0} \cdot \bm{v}_{1} = 0, \enspace & \bm{v}_{0} \perp \bm{v}_{1}.
    \end{cases}
\end{equation*}
The details of the inequality constraint optimization algorithm are provided in Alg.~\ref{alg:ineqConstAlg}.
\input{ineqConst}
\subsection{Cost Function Optimization Subroutine}
The cost function has the least priority in the optimization problem. The cost is reduced only if there is still any freedom or additional dimension left in the nullspace of the constraint Jacobian space. The cost function optimization algorithm is provided in Alg.~\ref{alg:costFuncAlg}.
\input{costFunc}
\subsection{Overall Framework and Termination Conditions}
The proposed overall framework combines Alg.~\ref{alg:eqConstAlg}-\ref{alg:costFuncAlg} along with some set of termination conditions. These conditions are iteration number, step tolerance, and cost tolerance, respectively. Each conditions are checked per outer iteration, that is one iteration of the overall framework altogether with Alg.~\ref{alg:eqConstAlg}-\ref{alg:costFuncAlg}. Similarly, the gradient descent iteration in each algorithm loop is called inner iteration.

The step tolerance checks the norm of the optimization parameter vector difference per outer iteration steps. If the parameter change is smaller than a given tolerance, the optimizer terminates. The same approach is also followed for the cost function return value check per outer iteration. The overall framework is provided in Alg.~\ref{alg:overallAlg}.
\input{overalFramework}
\section{Software Interface}
The numerical function optimization tool proposed in this paper is provided as a C++ dynamic library for community use. The software interface requires the users to write their optimization problems inside a given \textit{ProblemDescription} object template. There are four predefined functions, whose declaration and name cannot be changed, but only the content. These functions are \textit{costFunction}, \textit{equalityConstraintFunction}, \textit{inequalityConstraintFunction}, and \textit{interimFunction}, respectively. The first three are used for cost and constraint function iterations and calculations. These functions are repeatedly called every per inner iteration for numerical iterations and numerical gradient estimations. Even though the users are free to implement whatever they desire inside these functions' bodies, they should also consider the performance outcomes. An expensive computation inside these functions results in high computational cost. The \textit{interimFunction}, on the other hand, is called once per outer iteration and does not go into numerical gradient estimation. This function is placed to allow users to do less occasional parameter updates for performance considerations. For details and implementations, refer to ``NFO.hpp'' and ``ProblemDescription.hpp'' provided in \ref{github}.
\section{Numerical Results}
This section analyses five different problems with a parameter set given in Table.~\ref{parameterTab}. Three of these are numerical, and two are robotic system examples. The implementation codes for the first four examples are provided in \ref{github}. The optimization times correspond to a daily use desktop computer with AMD Ryzen 7 5800X CPU.
\begin{table}[t!] \centering \caption{List of given optimization parameters.}
\begin{tabular}{ccccc}
 \text{Parameter Name} & &&& \text{Value} \\ \cline{1-5} \rule{0pt}{3ex}
 \text{initial\_step\_length} & &&& $10^{-6}$ \\
 \text{step\_multiplier} & &&& $2$ \\
 \text{step\_tol} & &&& $10^{-4}$ \\
 \text{cost\_tol} & &&& $10^{-4}$
\end{tabular}
\label{parameterTab}
\end{table}
\subsection{Example 1: A Simple Convex Optimization Problem}
Assume the following optimization problem,
\begin{equation*}
    \begin{gathered}
        \min_{\bm{x}}{f(\bm{x}) = \sum_{k = 1}^{5}{(x_{k} - k)^{2}}} \\
        \text{such that} \\
        x_{1} + 5 = 0, \enspace x_{2} - 5 = 0, \\
        x_{3} + 3 < 0, \enspace x_{4} - 3 < 0, \enspace \bm{x}_{0} = \bm{0}
    \end{gathered}
\end{equation*}
where the unconstrained solution is obvious and equal to $[1,2,3,4,5]$. The constrained optimization solution is $\bm{x}^{*} = [-5,5,-3,3,5]^{\top}$. The first two parameters are equality constrained. Hence, the orthogonal projection of the cost function gradient onto the nullspace of the equality constraint Jacobian space results in zero in the direction of $x_{1}$ and $x_{2}$. The third and fourth parameters approach their unconstrained values as much as the inequality constraint allows. The fifth parameter, on the other hand, does not have any constraint and converges into its unconstrained optimal value. The optimization problem takes $8\mu s$ and is solved in 6 iterations.
\subsection{Example 2: Rosenbrock's Function}
Rosenbrock's function is a standard test function in optimization. Finding the minimum is a challenge for some algorithms because the function has a shallow minimum inside a deeply curved valley. Assume the following nonlinear constrained optimization problem with Rosenbrock's function,
\begin{equation*}
    \begin{gathered}
        \min_{\bm{x}}{f(\bm{x})} = 100(x_{2} - x_{1}^{2})^{2} + (1-x_{1})^{2} \\
        \text{such that} \\
        x_{1}^{2} + x_{2}^{2} \leq 1 \enspace \text{and} \enspace \bm{x}_{0} = [0,0].
    \end{gathered}
\end{equation*}
The minimum is found at $\bm{x}^{*} = [0.7864, 0.6177]^{\top}$ in 201 iterations which takes $35\mu s$.
\subsection{Example 3: Number 71 From the Hock-Schittkowsky Test Suite (\cite{HSref})}
A more challenging optimization problem can be found in the Hock-Schittkowsky test suite. Problem HS071 includes a high number of constraints that continuously disturb each other. For the given problem,
\begin{equation*}
    \begin{gathered}
        \min_{\bm{x}}{f(\bm{x}) = x_{1}x_{4}(x_{1} + x_{2} + x_{3}) + x_{3}} \\
        \text{such that} \\
        x_{1}x_{2}x_{3}x_{4} \geq 25 \\
        x_{1}^{2} + x_{2}^{2} + x_{3}^{2} + x_{4}^{2} = 40 \\
        1 \leq x_{1},x_{2},x_{3},x_{4} \leq 5 \\
        x_{0} = (1,5,5,1)
    \end{gathered}
\end{equation*}
the minimum is found at $\bm{x}^{*} \approx [1.00, 4.74, 3.82, 1.38]^{\top}$ in 100 iterations which takes $139\mu s$.
\subsection{Example 4: A 3-link Arm Configuration Optimization}
\begin{figure}[t!]
\centerline{\includegraphics[height=5cm]{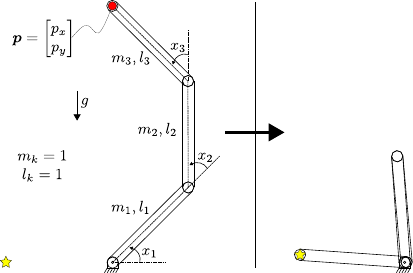}}
\caption{Left: A three-link planar robot arm system, where the red dot and yellow star represent the initial condition and the desired position, respectively. Right: The result of the optimization problem where the arm folds onto itself to minimize the static torque requirement while satisfying the equality constraint.}
\label{armFig}
\end{figure}

Assume a three-link planar robotic arm system as shown in Fig.~\ref{armFig}. The end effector position $\bm{p}(\bm{x})$ is a nonlinear function of optimization parameters:
\begin{equation*}
    \begin{bmatrix}
        p_{x} \\
        p_{y}
    \end{bmatrix}
    =
    \begin{bmatrix}
        \cos{(x_{1})} + \cos{(x_{1} + x_{2})} + \cos{(x_{1} + x_{2} + x_{3})} \\
        \sin{(x_{1})} + \sin{(x_{1} + x_{2})} + \sin{(x_{1} + x_{2} + x_{3})}
    \end{bmatrix}.
\end{equation*}
Similarly, the gravity vector of the system dynamics is given as
\begin{equation*}
\resizebox{0.99\hsize}{!}{$%
    \bm{\tau}_{g}(\bm{x}) = \begin{bmatrix}
        2.5 \cos{(x_{1})} + 1.5\cos{(x_{1} + x_{2})} + 0.5\cos{(x_{1} + x_{2} + x_{3})} \\
        1.5\cos{(x_{1} + x_{2})} + 0.5\cos{(x_{1} + x_{2} + x_{3})} \\
        0.5\cos{(x_{1} + x_{2} + x_{3})} \\
    \end{bmatrix}g.
$}
\end{equation*}
The gravity vector is equivalent to the amount of joint torques required to hold the robot in a static configuration. Hence, an optimization problem can be defined such that the end effector is placed at a desired position with a configuration that requires a minimum amount of torque to remain stationary. The problem formulation follows
\begin{equation*}
    \begin{gathered}
        \min_{\bm{x}}{f(\bm{x}) = \bm{\tau}_{g}^{\top}\bm{\tau}_{g}} \\
        \text {such that} \\
        p_{x}(\bm{x}) = -1.0, \enspace p_{y}(\bm{x}) = 0.0, \enspace and \enspace \bm{x}_{0} = \begin{bmatrix} \pi/4 & \pi/4 & \pi/4 \end{bmatrix}^{\top}.
    \end{gathered}
\end{equation*}
The optimized robot configuration is shown in Fig.~\ref{armFig}. The minimum is found at $\bm{x}^{*} \approx [1.647,3.141,-1.647]^{\top}$ in 8 iterations which takes $42\mu s$.
\subsection{Humanoid Robot Posture Optimization}
\begin{figure}[b!]
\centerline{\includegraphics[width=\columnwidth]{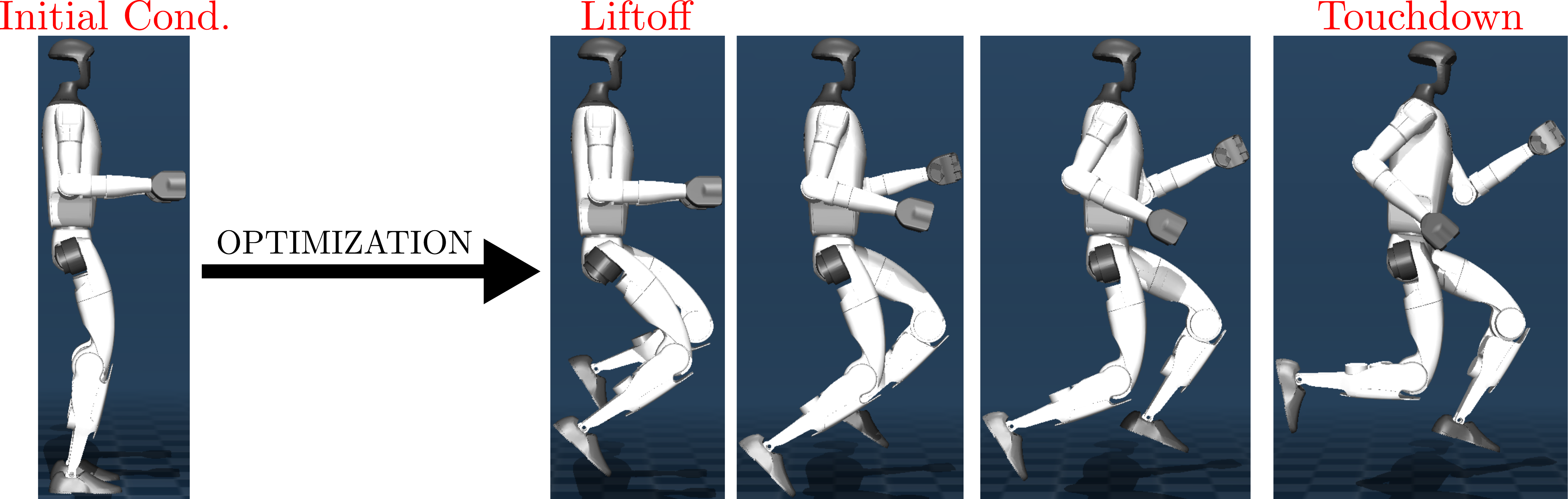}}
\caption{The snapshots of the optimized trajectory.}
\label{g1Fig}
\end{figure}

A high dimensional nonlinear robotic optimization problem that is solved with the proposed function optimization framework can be found in \cite{sovukluk2025}. The optimization problem covers finding a set of joint trajectories which results in a proper running motion. More specifically, the optimization problem covers how a running humanoid robot should swing its limbs during the flight phases such that at the next landing, the feet are at desired locations and the torso is kept upright. The snapshots of the optimized trajectories are shown in Fig.~\ref{g1Fig}. The optimization problem includes 32 optimization parameters along with 14 nonlinear constraints and takes $1.92 ms$ to solve (\cite{sovukluk2025}). In the implementation, the \textit{interimFunction} is used effectively for performance considerations, such that, during the trajectory optimization, the system dynamics are updated at the beginning of every outer iteration rather than the inner iterations.
\section{Conclusion}
This paper proposes a numerical function optimization framework designed to solve constrained nonlinear robotic problems in real-time. The proposed framework does not require any analytical representation of the problem and can also be referred as a constrained black-box optimization tool. The method combines first-order gradient-based line search algorithms with constraint prioritization through nullspace projections of constraint Jacobian space. Consequently, it usually has a slower convergence rate per iteration as the Hessian is ignored. On the other hand, the computational cost per iteration is much less as it only requires basic numerical gradient calculations. The capability of the framework is proven through numerous complex numerical and robotic problems ranging from a simple robotic arm to a complex humanoid robot.
\bibliography{ifacconf}
\end{document}

%% file: eqConst.tex
\begin{algorithm} [t!]
\caption{Optimizing for equality constraints.}\label{alg:eqConstAlg}
\begin{algorithmic} [1]
\Require $\bm{x}\in\mathbb{R}^{n}$, $n_{\text{ce}} \geq 0$
\State $\bm{J}_{\text{eq}}.\text{empty()}$
\State $n_{\text{ac}} \gets 0$
\For{($k \gets 1;\enspace k \leq n_{\text{ec}};\enspace \text{++k}$)}
    \State $n_{\text{ac}} = n_{\text{ac}} + 1$
    \State $\text{cost} = g_{\text{ec}}(\bm{x})$
    \State $\nabla g_{\text{ec},k} = \text{sign(cost)}\times\text{numericalGradient}(g_{\text{ec},k}(\bm{x}))$
    \If{$||\nabla g_{\text{ec},k}|| \approx 0$}
        \State $\text{Continue}$ \Comment{Zero gradient, skip to the next.}
    \EndIf
    \State $\bm{J}_{\text{eq}}.\text{rowAppend}(\nabla g_{\text{ec},k}^{\top})$
    \If{$|\text{cost}| \approx 0$}
        \State $\text{Continue}$ \Comment{No need to solve, skip to the next.}
    \EndIf
    \If{$n_{\text{ac}} > 1$} \Comment{Perform projection if necessary.}
        \State $\nabla \bm{g}^{*} = \text{proj}_{N(\bm{J}_{\text{eq}})}\nabla g_{\text{ec},k}$
    \Else
        \State $\nabla \bm{g}^{*} = \nabla g_{\text{ec},1}$
    \EndIf
    \If{$||\nabla \bm{g}^{*}|| \approx 0$}
        \State $\text{Continue}$ \Comment{Empty nullspace, skip to the next.}
    \EndIf
    \State $\text{stepLength} \gets \text{initial\_step\_length}$
    \While{true}
        \State $\bm{x}^{*} = \bm{x} - \text{stepLength}\times \nabla\bm{g}^{*}$
        \State $\text{cost}^{*} = g_{\text{ec},k}(\bm{x}^{*})$
        \If{$|\text{cost}^{*}| > |\text{cost}|$}
            \State \text{Break} \Comment{Minimal point has reached.}
        \EndIf
        \If{$\text{sign}(\text{cost}^{*}) \neq \text{sign}(\text{cost})$}
            \Statex \Comment{Zero crossing, reverse gradient direction.}
            \State $\nabla\bm{g}^{*} = -\nabla\bm{g}^{*}$
        \EndIf
        \State $\bm{x} \gets \bm{x}^{*}$
        \State $\text{stepLength} = \text{stepLength} \times \text{step\_multiplier}$
    \EndWhile
\EndFor
\end{algorithmic}
\end{algorithm}

%% file: ineqConst.tex
\begin{algorithm} [t!]
\caption{Optimizing for inequality constraints.}\label{alg:ineqConstAlg}
\begin{algorithmic} [1]
\Require $\bm{x}\in\mathbb{R}^{n}$ \Comment{From equality constraint algorithm.}
\Require $\bm{J_{\text{eq}}}$, $n_{\text{ce}}$ \Comment{From equality constraint algorithm.}
\State $\bm{J}_{\text{ineq}}.\text{empty()}$
\For{($k \gets 1;\enspace k \leq n_{\text{ic}};\enspace \text{++k}$)}
    \State $\bm{J}_{\text{active}} \gets \bm{J}_{\text{eq}}$
    \State $\text{cost} = g_{\text{ic}}(\bm{x})$
    \If{$\text{cost} < 0$}
        \State $\text{Continue}$ \Comment{No need to solve, skip to the next.}
    \EndIf
    \State $\nabla g_{\text{ic},k} = \text{numericalGradient}(g_{\text{ic},k}(\bm{x}))$
    \If{$||\nabla g_{\text{ic},k}|| \approx 0$}
        \State $\text{Continue}$ \Comment{Zero gradient, skip to the next.}
    \EndIf
    \State $n_{\text{ac}} = n_{\text{ac}} + 1$
    \State $\bm{J}_{\text{ineq}}.\text{rowAppend}(\nabla g_{\text{ic},k}^{\top})$
    \Statex \Comment{Check previous active constraints' grad directions.}
    \If{$\text{rowNumber}(\bm{J}_{\text{ineq}}) > 1$}
        \For{($i \gets 1; i \leq \text{rowNumber}(\bm{J}_{\text{ineq}}); \text{++i}$)}
            \If{$\bm{J}_{\text{ineq}}.\text{row}(i) \cdot \nabla g_{\text{ic},k} < 0$}
                \State $\bm{J}_{\text{active}}.\text{rowAppend}(\bm{J}_{\text{ineq}}.\text{row}(i))$
            \EndIf
        \EndFor
    \EndIf
    \Statex \Comment{Then initiate the projection operation.}
    \If{$n_{\text{ac}} > 1$}
        \State $\nabla \bm{g}^{*} = \text{proj}_{N(\bm{J}_{\text{active}})}\nabla g_{\text{ic},k}$
    \Else
        \State $\nabla \bm{g}^{*} = \nabla g_{\text{ec},1}$
    \EndIf
    \If{$||\nabla \bm{g}^{*}|| \approx 0$}
        \State $\text{Continue}$ \Comment{Empty nullspace, skip to the next.}
    \EndIf
    \State $\text{stepLength} \gets \text{initial\_step\_length}$
    \While{true}
        \State $\bm{x}^{*} = \bm{x} - \text{stepLength}\times \nabla\bm{g}^{*}$
        \State $\text{cost}^{*} = g_{\text{ic},k}(\bm{x}^{*})$
        \If{$\text{cost}^{*} < 0$}
            \State \text{Break} \Comment{Zero crossing has reached.}
        \EndIf
        \State $\bm{x} \gets \bm{x}^{*}$
        \State $\text{stepLength} = \text{stepLength} \times \text{step\_multiplier}$
    \EndWhile
\EndFor
\end{algorithmic}
\end{algorithm}

%% file: costFunc.tex
\begin{algorithm} [t!]
\caption{Optimizing for the cost function.}\label{alg:costFuncAlg}
\begin{algorithmic} [1]
\Require $\bm{x}\in\mathbb{R}^{n}$ \Comment{From inequality constraint alg.}
\Require $\bm{J}_{\text{eq}}$, $n_{\text{ce}}$ \Comment{From inequality constraint alg.}
\Require $\bm{J}_{\text{ineq}}$ \Comment{From inequality constraint alg.}
\State $\bm{J}_{\text{active}} \gets \bm{J}_{\text{eq}}$
\State $\text{cost} = f(\bm{x})$
\State $\nabla f = \text{numericalGradient}(f(\bm{x}))$
\If{$||\nabla f|| \approx 0$}
    \State $\text{Continue}$ \Comment{Zero gradient, skip.}
\EndIf
\Statex \Comment{Check previous active constraints' grad directions.}
\If{$\text{rowNumber}(\bm{J}_{\text{ineq}}) > 0$}
    \For{($i \gets 1; i \leq \text{rowNumber}(\bm{J}_{\text{ineq}}); \text{++i}$)}
        \If{$\bm{J}_{\text{ineq}}.\text{row}(i) \cdot \nabla f < 0$}
            \State $\bm{J}_{\text{active}}.\text{rowAppend}(\bm{J}_{\text{ineq}}.\text{row}(i))$
        \EndIf
    \EndFor
\EndIf
\Statex \Comment{Initiate the projection operation if necessary.}
\If{$n_{\text{ac}} > 0$}
    \State $\nabla \bm{f}^{*} = \text{proj}_{N(\bm{J}_{\text{active}})}\nabla f$
    \If{$||\nabla f^{*}|| \approx 0$}
        \State $\text{Continue}$ \Comment{Empty nullspace, skip.}
    \EndIf
\EndIf
\State $\text{stepLength} \gets \text{initial\_step\_length}$
\While{true}
    \State $\bm{x}^{*} = \bm{x} - \text{stepLength}\times \nabla\bm{f}^{*}$
    \State $\text{cost}^{*} =f(\bm{x}^{*})$
    \If{$\text{cost}^{*} > \text{cost}$}
        \State \text{Break} \Comment{Minimal point has reached.}
    \EndIf
    \Statex \Comment{Check if $\bm{x}^{*}$ violates any inequality constraint.}
    \For{($i \gets 1; i \leq n_\text{ic}; i=i+1$)}
        \If{$g_{\text{ineq}}(\bm{x}^{*}) > 0$}
            \State \text{Break the while loop.}
        \EndIf
    \EndFor
    \State $\bm{x} \gets \bm{x}^{*}$
    \State $\text{stepLength} = \text{stepLength} \times \text{step\_multiplier}$
\EndWhile
\end{algorithmic}
\end{algorithm}

%% file: overalFramework.tex
\begin{algorithm} [t!]
\caption{The overall optimization framework.}\label{alg:overallAlg}
\begin{algorithmic} [1]
\Require $\bm{x}_{0}$ \Comment{Initial condition.}
\Require \text{max\_iter}
\Require \text{step\_tol}
\Require \text{cost\_tol}
\State $\bm{x} \gets \bm{x}_{0}$
\For {$(\text{iter} \gets 1;\enspace\text{iter} \leq \text{max\_iter};\enspace\text{++iter})$}
    \State $\bm{x}_{0} \gets \bm{x}$
    \State \text{Call Alg.~\ref{alg:eqConstAlg}}
    \State \text{Call Alg.~\ref{alg:ineqConstAlg}}
    \State $\text{cost}_{0} \gets f(\bm{x})$
    \State \text{Call Alg.~\ref{alg:costFuncAlg}}

    \If {$||\bm{x} - \bm{x}_{0}|| < \text{step\_tol}$}
        \State \text{Exit}
    \EndIf
    \If {$|\text{cost} - \text{cost}_{0}| < \text{cost\_tol}$}
        \State \text{Exit}
    \EndIf
\EndFor
\end{algorithmic}
\end{algorithm}

%% file: ifacconf.bbl
\begin{thebibliography}{25}
\providecommand{\natexlab}[1]{#1}
\providecommand{\url}[1]{\texttt{#1}}
\providecommand{\urlprefix}{URL }
\expandafter\ifx\csname urlstyle\endcsname\relax
  \providecommand{\doi}[1]{doi:\discretionary{}{}{}#1}\else
  \providecommand{\doi}{doi:\discretionary{}{}{}\begingroup \urlstyle{rm}\Url}\fi

\bibitem[{Andersson et~al.(2019)Andersson, Gillis, Horn, Rawlings, and Diehl}]{casadiRef}
Andersson, J.A., Gillis, J., Horn, G., Rawlings, J.B., and Diehl, M. (2019).
\newblock Casadi: a software framework for nonlinear optimization and optimal control.
\newblock \emph{Mathematical Programming Computation}, 11, 1--36.

\bibitem[{Bambade et~al.(2022)Bambade, El-Kazdadi, Taylor, and Carpentier}]{proxqpRef}
Bambade, A., El-Kazdadi, S., Taylor, A., and Carpentier, J. (2022).
\newblock Prox-qp: Yet another quadratic programming solver for robotics and beyond.
\newblock In \emph{RSS 2022-Robotics: Science and Systems}.

\bibitem[{Bertsekas(1997)}]{nonlinearProg}
Bertsekas, D.P. (1997).
\newblock Nonlinear programming.
\newblock \emph{Journal of the Operational Research Society}, 48(3), 334--334.

\bibitem[{Carpentier et~al.(2019)Carpentier, Saurel, Buondonno, Mirabel, Lamiraux, Stasse, and Mansard}]{carpentier2019pinocchio}
Carpentier, J., Saurel, G., Buondonno, G., Mirabel, J., Lamiraux, F., Stasse, O., and Mansard, N. (2019).
\newblock The pinocchio c++ library -- a fast and flexible implementation of rigid body dynamics algorithms and their analytical derivatives.
\newblock In \emph{IEEE International Symposium on System Integrations (SII)}.

\bibitem[{Carpentier et~al.(2015--2021)Carpentier, Valenza, Mansard et~al.}]{pinocchioweb}
Carpentier, J., Valenza, F., Mansard, N., et~al. (2015--2021).
\newblock Pinocchio: fast forward and inverse dynamics for poly-articulated systems.
\newblock https://stack-of-tasks.github.io/pinocchio.

\bibitem[{Dai et~al.(2014)Dai, Valenzuela, and Tedrake}]{wholeBodyMotionRef}
Dai, H., Valenzuela, A., and Tedrake, R. (2014).
\newblock Whole-body motion planning with centroidal dynamics and full kinematics.
\newblock In \emph{2014 IEEE-RAS International Conference on Humanoid Robots}, 295--302.
\newblock \doi{10.1109/HUMANOIDS.2014.7041375}.

\bibitem[{Featherstone(2014)}]{featherstone2014rigid}
Featherstone, R. (2014).
\newblock \emph{Rigid body dynamics algorithms}.
\newblock Springer.

\bibitem[{Felis(2016)}]{RBDL}
Felis, M.L. (2016).
\newblock Rbdl: an efficient rigid-body dynamics library using recursive algorithms.
\newblock \emph{Autonomous Robots}, 1--17.
\newblock \doi{10.1007/s10514-016-9574-0}.

\bibitem[{Gasparetto et~al.(2015)Gasparetto, Boscariol, Lanzutti, and Vidoni}]{trajOptim}
Gasparetto, A., Boscariol, P., Lanzutti, A., and Vidoni, R. (2015).
\newblock Path planning and trajectory planning algorithms: A general overview.
\newblock \emph{Motion and operation planning of robotic systems: Background and practical approaches}, 3--27.

\bibitem[{Haldane et~al.(2017)Haldane, Yim, and Fearing}]{saltoRef}
Haldane, D.W., Yim, J.K., and Fearing, R.S. (2017).
\newblock Repetitive extreme-acceleration (14-g) spatial jumping with salto-1p.
\newblock In \emph{2017 IEEE/RSJ International Conference on Intelligent Robots and Systems (IROS)}, 3345--3351. IEEE.

\bibitem[{Hock and Schittkowski(1980)}]{HSref}
Hock, W. and Schittkowski, K. (1980).
\newblock Test examples for nonlinear programming codes.
\newblock \emph{Journal of optimization theory and applications}, 30, 127--129.

\bibitem[{Khatib et~al.(2022)Khatib, Jorda, Park, Sentis, and Chung}]{khatipRef}
Khatib, O., Jorda, M., Park, J., Sentis, L., and Chung, S.Y. (2022).
\newblock Constraint-consistent task-oriented whole-body robot formulation: Task, posture, constraints, multiple contacts, and balance.
\newblock \emph{The International Journal of Robotics Research}, 41(13-14), 1079--1098.
\newblock \doi{10.1177/02783649221120029}.

\bibitem[{Lewis et~al.(2012)Lewis, Vrabie, and Syrmos}]{optimalRef}
Lewis, F.L., Vrabie, D., and Syrmos, V.L. (2012).
\newblock \emph{Optimal control}.
\newblock John Wiley \& Sons.

\bibitem[{Mei et~al.(2004)Mei, Lu, Hu, and Lee}]{energyRef1}
Mei, Y., Lu, Y.H., Hu, Y., and Lee, C. (2004).
\newblock Energy-efficient motion planning for mobile robots.
\newblock In \emph{IEEE International Conference on Robotics and Automation, 2004. Proceedings. ICRA '04. 2004}, volume~5, 4344--4349 Vol.5.
\newblock \doi{10.1109/ROBOT.2004.1302401}.

\bibitem[{Nocedal and Wright(1999)}]{numericalOptim}
Nocedal, J. and Wright, S.J. (1999).
\newblock \emph{Numerical optimization}.
\newblock Springer.

\bibitem[{Paes et~al.(2014)Paes, Dewulf, Elst, Kellens, and Slaets}]{energyRef2}
Paes, K., Dewulf, W., Elst, K.V., Kellens, K., and Slaets, P. (2014).
\newblock Energy efficient trajectories for an industrial abb robot.
\newblock \emph{Procedia CIRP}, 15, 105--110.
\newblock \doi{https://doi.org/10.1016/j.procir.2014.06.043}.
\newblock 21st CIRP Conference on Life Cycle Engineering.

\bibitem[{Pandala et~al.(2019)Pandala, Ding, and Park}]{qpswiftRef}
Pandala, A.G., Ding, Y., and Park, H.W. (2019).
\newblock qpswift: A real-time sparse quadratic program solver for robotic applications.
\newblock \emph{IEEE Robotics and Automation Letters}, 4(4), 3355--3362.

\bibitem[{Sovukluk et~al.(2023{\natexlab{a}})Sovukluk, Englsberger, and Ott}]{sovukluk1}
Sovukluk, S., Englsberger, J., and Ott, C. (2023{\natexlab{a}}).
\newblock Whole body control formulation for humanoid robots with closed/parallel kinematic chains: Kangaroo case study.
\newblock In \emph{2023 IEEE/RSJ International Conference on Intelligent Robots and Systems (IROS)}, 10390--10396.
\newblock \doi{10.1109/IROS55552.2023.10341391}.

\bibitem[{Sovukluk et~al.(2024)Sovukluk, Englsberger, and Ott}]{sovukluk2}
Sovukluk, S., Englsberger, J., and Ott, C. (2024).
\newblock Highly maneuverable humanoid running via 3d slip+foot dynamics.
\newblock \emph{IEEE Robotics and Automation Letters}, 9(2), 1131--1138.
\newblock \doi{10.1109/LRA.2023.3342668}.

\bibitem[{Sovukluk et~al.(2023{\natexlab{b}})Sovukluk, Ott, and Ankaralı}]{sovukluk2023cascaded}
Sovukluk, S., Ott, C., and Ankaralı, M.M. (2023{\natexlab{b}}).
\newblock Cascaded model predictive control of underactuated bipedal walking with impact and friction considerations.
\newblock In \emph{2023 IEEE-RAS 22nd International Conference on Humanoid Robots (Humanoids)}, 1--8.
\newblock \doi{10.1109/Humanoids57100.2023.10375153}.

\bibitem[{Sovukluk et~al.(2025)Sovukluk, Schuller, Englsberger, and Ott}]{sovukluk2025}
Sovukluk, S., Schuller, R., Englsberger, J., and Ott, C. (2025).
\newblock Realtime limb trajectory optimization for humanoid running through centroidal angular momentum dynamics.
\newblock \doi{10.48550/arXiv.2501.17351}.

\bibitem[{Stellato et~al.(2020)Stellato, Banjac, Goulart, Bemporad, and Boyd}]{osqpRef}
Stellato, B., Banjac, G., Goulart, P., Bemporad, A., and Boyd, S. (2020).
\newblock {OSQP}: an operator splitting solver for quadratic programs.
\newblock \emph{Mathematical Programming Computation}, 12(4), 637--672.
\newblock \doi{10.1007/s12532-020-00179-2}.

\bibitem[{Todorov et~al.(2012)Todorov, Erez, and Tassa}]{todorov2012mujoco}
Todorov, E., Erez, T., and Tassa, Y. (2012).
\newblock Mujoco: A physics engine for model-based control.
\newblock In \emph{2012 IEEE/RSJ International Conference on Intelligent Robots and Systems}, 5026--5033. IEEE.
\newblock \doi{10.1109/IROS.2012.6386109}.

\bibitem[{W{\"a}chter and Biegler(2006)}]{ipopt}
W{\"a}chter, A. and Biegler, L.T. (2006).
\newblock On the implementation of an interior-point filter line-search algorithm for large-scale nonlinear programming.
\newblock \emph{Mathematical programming}, 106, 25--57.

\bibitem[{Westervelt et~al.(2018)Westervelt, Grizzle, Chevallereau, Choi, and Morris}]{westervelt2018feedback}
Westervelt, E.R., Grizzle, J.W., Chevallereau, C., Choi, J.H., and Morris, B. (2018).
\newblock \emph{Feedback control of dynamic bipedal robot locomotion}.
\newblock CRC press.

\end{thebibliography}
